\documentclass[10pt]{article}
\usepackage[margin=1.05in]{geometry}
\usepackage{amsmath,amssymb,bm}
\usepackage{booktabs}
\usepackage{graphicx}
\usepackage{xcolor}
\usepackage[colorlinks=true,linkcolor=blue,citecolor=blue,urlcolor=blue]{hyperref}
\usepackage{lmodern}
\usepackage{microtype}
\usepackage[numbers]{natbib}

\newcommand{\staFIVEnhTHREEbmpTHREETWO}{0.21}

\newcommand{\staFIVEnhTHREEbmpFIVEONETWO}{0.20}

\newcommand{\staFIVEnhTHREEbppTHREETWO}{0.21}

\newcommand{\staFIVEnhTHREEbppFIVEONETWO}{0.19}

\newcommand{\staFIVEnhFOURbmpTHREETWO}{1.00}

\newcommand{\staFIVEnhFOURbmpFIVEONETWO}{1.00}

\newcommand{\staFIVEnhFOURbpbzFIVEONETWO}{1.00}
\newcommand{\staFIVEnhFOURbpbzgen}{4}

\newcommand{\staFIVEnhFOURbppTHREETWO}{1.00}

\newcommand{\staFIVEnhFOURbppFIVEONETWO}{0.22}

\newcommand{\stbfitseeds}{49}
\newcommand{\stbfitsurvive}{49}

\newcommand{\stminnhaFIVE}{4}

\newcommand{\stminnhparity}{1}
\newcommand{\stminnhsFOUR}{3}
\newcommand{\stminnhsFIVE}{4}
\newcommand{\stmodTHREEnhTWObmpTHREETWO}{1.00}

\newcommand{\stmodTHREEnhTWObmpFIVEONETWO}{0.34}

\newcommand{\stmodTHREEnhTWObpbzFIVEONETWO}{0.34}
\newcommand{\stmodTHREEnhTWObpbzgen}{2}

\newcommand{\stmodTHREEnhTWObppTHREETWO}{1.00}

\newcommand{\stmodTHREEnhTWObppFIVEONETWO}{0.33}

\newcommand{\stmodTHREEnhFOURbmpFIVEONETWO}{0.00}

\newcommand{\stparitynhONEbmXpFIVEONETWO}{1.00}

\newcommand{\stparitynhONEbmpTHREETWO}{1.00}

\newcommand{\stparitynhONEbmpFIVEONETWO}{1.00}

\newcommand{\stparitynhONEbpXbzFIVEONETWO}{1.00}

\newcommand{\stparitynhONEbpXpFIVEONETWO}{0.50}

\newcommand{\stparitynhONEbpXusageend}{0.41}

\newcommand{\stparitynhONEbpbzFIVEONETWO}{1.00}
\newcommand{\stparitynhONEbpbzgen}{5}

\newcommand{\stparitynhONEbppTHREETWO}{1.00}

\newcommand{\stparitynhONEbppFIVEONETWO}{0.50}

\newcommand{\stparitynhTWObpbzgen}{1}

\newcommand{\stparitynhFOURbpbzgen}{2}

\newcommand{\stsFOURnhTWObmpTHREETWO}{0.27}

\newcommand{\stsFOURnhTWObmpFIVEONETWO}{0.24}

\newcommand{\stsFOURnhTWObppTHREETWO}{0.26}

\newcommand{\stsFOURnhTWObppFIVEONETWO}{0.24}

\newcommand{\stsFOURnhTHREEbmpTHREETWO}{1.00}

\newcommand{\stsFOURnhTHREEbmpFIVEONETWO}{1.00}

\newcommand{\stsFOURnhTHREEbpbzFIVEONETWO}{0.99}
\newcommand{\stsFOURnhTHREEbpbzgen}{4}

\newcommand{\stsFOURnhTHREEbppTHREETWO}{1.00}

\newcommand{\stsFOURnhTHREEbppFIVEONETWO}{0.26}

\newcommand{\stsFIVEnhTHREEbmpTHREETWO}{0.27}

\newcommand{\stsFIVEnhTHREEbmpFIVEONETWO}{0.20}

\newcommand{\stsFIVEnhTHREEbppTHREETWO}{0.28}

\newcommand{\stsFIVEnhTHREEbppFIVEONETWO}{0.19}

\newcommand{\stsFIVEnhFOURbmXpFIVEONETWO}{1.00}

\newcommand{\stsFIVEnhFOURbmpTHREETWO}{1.00}

\newcommand{\stsFIVEnhFOURbmpFIVEONETWO}{1.00}

\newcommand{\stsFIVEnhFOURbpXbzFIVEONETWO}{1.00}

\newcommand{\stsFIVEnhFOURbpXpFIVEONETWO}{0.23}

\newcommand{\stsFIVEnhFOURbpXusageend}{0.77}
\newcommand{\stsFIVEnhFOURbpbzTHREETWO}{1.00}
\newcommand{\stsFIVEnhFOURbpbzFIVEONETWO}{0.99}
\newcommand{\stsFIVEnhFOURbpbzgen}{5}

\newcommand{\stsFIVEnhFOURbppTHREETWO}{1.00}

\newcommand{\stsFIVEnhFOURbppFIVEONETWO}{0.20}

\newcommand{\bt}{b_t}

\title{The Automaton Underneath:\\ The Additive Input Pathway Is a Parasitic Attractor\\
for State Tracking in Householder Linear RNNs}
\author{Gunner Levi Howe\\\texttt{gunnerlevihowe@gmail.com}}
\date{July 2026}

\begin{document}
\maketitle

\begin{abstract}
Linear RNNs with input-dependent Householder-product transitions (DeltaNet/DeltaProduct-class) can
provably represent hard state-tracking automata, yet trained models notoriously fail to
length-generalize --- a gap recent work names ``expressivity vs.\ learnability'' and explicitly
attributes to optimization, without a causal account. We give one, in a pre-registered,
within-architecture causal ablation: the same model, with one term deleted --- the additive input
injection $\bt = W_b e_t$. With $\bt$, models fit length~32 perfectly and collapse out-of-distribution
on parity, $S_4$, $A_5$, and the non-solvable $S_5$ word problems (median position-512 accuracy
$\stsFIVEnhFOURbppFIVEONETWO$ on $S_5$). Without $\bt$ --- input acting \emph{only} through the
orthogonal transitions --- the same architecture at the same width learns the \emph{exact} automaton:
median accuracy $\stsFIVEnhFOURbmpFIVEONETWO$ at $16\times$ the training length, at every width admitted
by a representation law we state and test: the minimal number of Householder factors per token equals
the maximal reflection length of the task's generators in the representation the task format pins
(parity $1$, $S_4$ $3$, $A_5$ and $S_5$ $4$); below it, $-b$ models cannot fit at all, and the
contrast with DeltaProduct's $S_4/A_5$-at-$n_h{=}2$ (group-element classification, $SO(3)$
realization) shows the law is representation-relative rather than absolute. Two pre-registered
discriminators locate the mechanism. (i)~Initialized \emph{at} a verified-exact solution with
$W_b{=}0$, Adam grows the additive path (probe $\lVert W_b e\rVert$ $0\!\to\!\stsFIVEnhFOURbpXusageend$
on $S_5$) and pulls the model off the exact solution ($\stsFIVEnhFOURbpXpFIVEONETWO$ at position 512)
while $-b$ controls stay at $1.00$ --- the additive route is an attractor, not merely a basin around
random inits. (ii)~Our pre-registered prediction that the shortcut's in-domain fit \emph{lives on}
$\bt$ was wrong, and the registered kill criterion fired: zeroing $W_b$ at inference leaves in-domain
fit intact in all 49 fitting seeds --- and at the law-minimal width it \emph{restores exact length
generalization} (parity $5/5$, $S_5$ $5/5$, $S_4$ $4/5$, $A_5$ $4/5$ seeds). The additive pathway is
parasitic: a norm-inflation channel that cross-entropy feeds, which first destabilizes and then
conceals a correctly learned automaton. Boundaries are reported with the same discipline: above-law
widths restore in only $1$--$2/5$ seeds, and the cyclic mod-3 task drifts rather than locking even
without $\bt$ --- killing our own earlier ``transposition anchor'' hypothesis, since all-even-generator
$A_5$ locks exactly. All 202 runs, predictions, and kill criteria were pre-registered in-repository
before any grid run; every number regenerates from run artifacts.
\end{abstract}

\section{Introduction}
State tracking --- maintaining the running state of an automaton while reading a sequence --- is the
sharpest known separation between attention/parallel architectures and true recurrence
\citep{merrill2024illusion,deletang2023chomsky}, and the leading sub-quadratic architectures
(DeltaNet-class linear RNNs with input-dependent orthogonal or near-orthogonal transitions) were
designed in part to close it \citep{grazzi2025negative,siems2025deltaproduct}. A striking 2025--2026
literature documents that they close it \emph{in principle but not in training}: models that provably
contain exact solutions fit the training length and collapse beyond it
\citep{shakerinava2026diagonal,mishra2026m2rnn,chung2026error,deletang2023chomsky}.
\citet{shakerinava2026diagonal} name the gap --- ``the main bottleneck \ldots is not expressivity but
optimization'' --- demonstrate it survives near-solution initialization in diagonal SSMs, and call for
a loss-landscape account. Cross-architecture evidence points at the input pathway:
\citet{terzic2025sdssm} report that diagonal selective SSMs \emph{without} the additive $B$ matrix
length-generalize better on commutative automata; \citet{ebrahimi2025bilinear} show pure bilinear
(non-affine) RNNs length-generalize on parity and random permutation FSMs where affine variants fail;
and injection-free orthogonal-transition architectures extrapolate spectacularly
\citep{sung2026holonomic,lee2026falsifier}. But none of these is a \emph{causal} test: each compares
different architectures (or removes capacity along with the pathway, as in diagonal SSMs, which lose
non-commutative expressivity with $B$), none explains \emph{why} gradient descent prefers the failing
solution, and PD-SSM's exact $S_5$ extrapolation \emph{with} an additive term \citep{terzic2025pdssm}
falsifies any architecture-universal reading.

This note supplies the missing experiment, pre-registered end-to-end. In a fixed
DeltaProduct-style Householder linear RNN --- where the orthogonal transitions alone are expressive
enough for every task studied, so deleting the additive injection $\bt$ changes \emph{what SGD finds}
but not \emph{what the model can represent} --- we delete exactly one term and observe, across five
seeds per cell:

\begin{enumerate}\itemsep2pt
\item \textbf{The flip} (\S\ref{sec:flip}): $+b$ models fit length 32 and collapse
  (shortcut); $-b$ models learn the exact automaton to length 512 ($16\times$), including the
  non-solvable $S_5$, at matched width, budget, and data.
\item \textbf{A representation law with a necessity side} (\S\ref{sec:law}): the minimal $n_h$ that
  length-generalizes equals the maximal generator reflection length $\operatorname{rank}(I-P)$ in the
  representation pinned by the task format (parity 1, $S_4$ 3, $A_5$/$S_5$ 4); every below-law $-b$
  cell is no-fit. DeltaProduct's $S_4/A_5$ at $n_h{=}2$ under group-element classification
  \citep{siems2025deltaproduct} is thereby reconciled, not contradicted: the law is
  representation-relative, and the task format selects the representation.
\item \textbf{The mechanism} (\S\ref{sec:mech}): from a verified-exact initialization with $W_b{=}0$,
  training \emph{grows} the additive path and destroys length generalization while $-b$ controls are
  stable --- and our registered prediction that the fit routes through $\bt$ was falsified by its own
  kill criterion: the additive path is \emph{parasitic}. At law-minimal width the exact automaton is
  learned \emph{underneath} it, and zeroing $W_b$ at inference --- no retraining --- restores exact
  generalization in $18/20$ seeds across parity/$S_4$/$S_5$/$A_5$.
\end{enumerate}

\section{Setup}\label{sec:setup}
\textbf{Model.} A single recurrent layer with input-dependent Householder-product transitions
\citep{siems2025deltaproduct}: $h_t = R_{n_h}(x_t)\cdots R_1(x_t)\,h_{t-1} + \bt$, where each
$R_i(x_t)=I-2u_i u_i^\top$ ($u_i$ a learned, normalized, input-dependent vector), $d{=}112$,
$\bt=W_b e_t$ is the standard additive injection, and a per-step \emph{linear} readout predicts the
automaton state. The $-b$ ablation deletes $\bt$ only: input then acts exclusively by selecting
orthogonal transitions --- the literal automaton form. Training: full 32-length sequences, 40 epochs,
Adam, identical budget in every cell (Appendix~\ref{app:repro}); no cell-specific tuning anywhere.

\textbf{Tasks.} Per-step tracking of a designated element's position under generator sequences
(vocabulary = 2 generators): parity ($\mathbb{Z}_2$), mod-3 ($\mathbb{Z}_3$), $S_4$
(transposition + 4-cycle), $A_5$ (3-cycle + 5-cycle; all-even), $S_5$ (transposition + 5-cycle;
non-solvable). Group closures and generator reflection lengths are asserted programmatically at
runner start. Train $L{=}32$, evaluate per-position to $L{=}512$; \textbf{GEN} = median position-512
accuracy ${>}0.9$; \textbf{shortcut} = fits position 32 (${>}0.9$) but ${<}0.6$ at 512;
\textbf{no-fit} otherwise. Five seeds per primary cell; medians with $[\min,\max]$.

\textbf{Pre-registration.} Grid, predictions, kill criteria, and decision rules were committed to the
repository \emph{before any grid run existed} (commit \texttt{2cf13e5}), with verdicts recorded
against them unchanged (\texttt{84e21c8}); the one exploratory antecedent (a 3-seed June pilot of the
parity/$S_5$ flip) is disclosed as such, and all its cells were re-run from scratch under this
harness. Every number in this paper is a macro regenerated from the 202 run artifacts and
byte-verified (Appendix~\ref{app:repro}).

\section{One deleted term flips shortcut to exact automaton}\label{sec:flip}
\begin{figure}[t]
\centering\includegraphics[width=.98\linewidth]{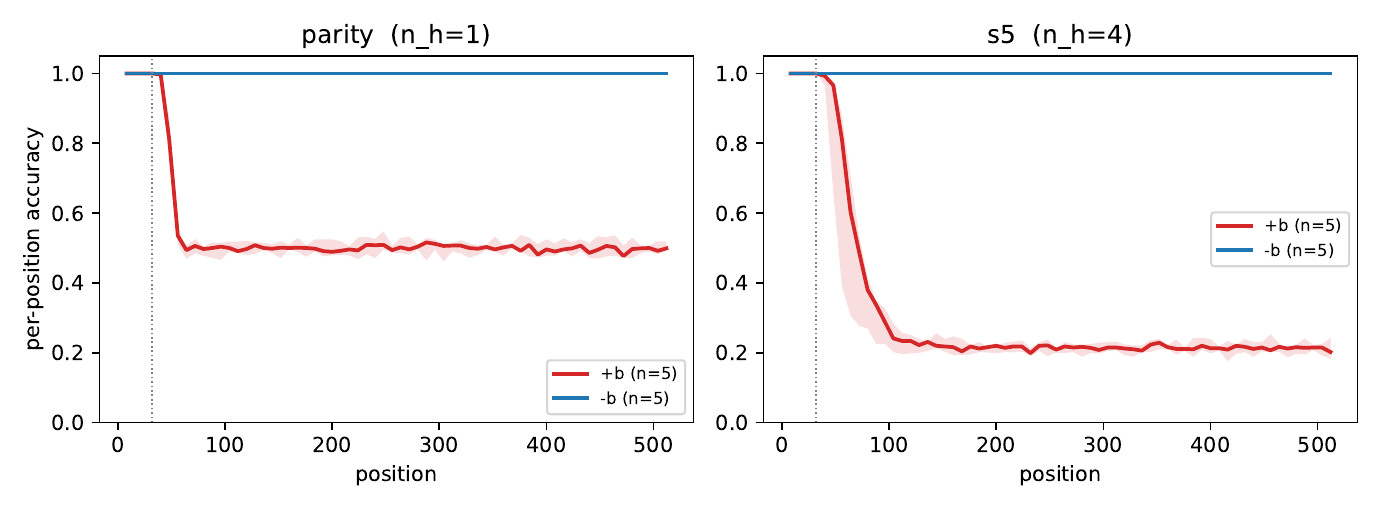}
\caption{Per-position accuracy, median (band = seed min--max), train $L{=}32$ (dotted). The same
architecture at the same width: $+b$ (red) fits and collapses; $-b$ (blue) is exact to $16\times$ the
training length, including non-solvable $S_5$.}
\label{fig:flip}
\end{figure}

Table~\ref{tab:main} is the paper's central result; Figure~\ref{fig:flip} shows the per-position
curves. With the additive path, every task that fits at all follows the shortcut profile: perfect at
the training length, collapsed by $2$--$4\times$ beyond it ($S_5$: $\stsFIVEnhFOURbppTHREETWO$ at
position 32 vs.\ $\stsFIVEnhFOURbppFIVEONETWO$ at 512). Deleting $\bt$ flips every such cell at or above
the law width to exact generalization: parity $\stparitynhONEbmpFIVEONETWO$, $S_4$
$\stsFOURnhTHREEbmpFIVEONETWO$, $A_5$ $\staFIVEnhFOURbmpFIVEONETWO$, $S_5$
$\stsFIVEnhFOURbmpFIVEONETWO$ at position 512 (medians of 5; $S_5$ and parity $[1.00,1.00]$). The GRU
reference generalizes everywhere (as known); the negative-eigenvalue diagonal model cannot fit $S_5$
(no order-5 real eigenvalue), reproducing the expressivity wall the Householder parameterization
exists to cross.

\begin{table}[t]
\centering\small
\begin{tabular}{llcccccc}
\toprule
& & \multicolumn{2}{c}{$+b$ (median pos-32 / pos-512)} & \multicolumn{2}{c}{$-b$ (pos-32 / pos-512)} &
\multicolumn{2}{c}{$+b$, $W_b{:=}0$ at inference} \\
task & $n_h$ & fit & OOD & fit & OOD & pos-512 & seeds GEN \\
\midrule
parity & 1 & $\stparitynhONEbppTHREETWO$ & $\stparitynhONEbppFIVEONETWO$ & $\stparitynhONEbmpTHREETWO$ & $\stparitynhONEbmpFIVEONETWO$ & $\stparitynhONEbpbzFIVEONETWO$ & $\stparitynhONEbpbzgen$/5 \\
mod-3 & 2 & $\stmodTHREEnhTWObppTHREETWO$ & $\stmodTHREEnhTWObppFIVEONETWO$ & $\stmodTHREEnhTWObmpTHREETWO$ & $\stmodTHREEnhTWObmpFIVEONETWO$ & $\stmodTHREEnhTWObpbzFIVEONETWO$ & $\stmodTHREEnhTWObpbzgen$/5 \\
$S_4$ & 2 & $\stsFOURnhTWObppTHREETWO$ & $\stsFOURnhTWObppFIVEONETWO$ & $\stsFOURnhTWObmpTHREETWO$ & $\stsFOURnhTWObmpFIVEONETWO$ & --- & --- \\
$S_4$ & 3 & $\stsFOURnhTHREEbppTHREETWO$ & $\stsFOURnhTHREEbppFIVEONETWO$ & $\stsFOURnhTHREEbmpTHREETWO$ & $\stsFOURnhTHREEbmpFIVEONETWO$ & $\stsFOURnhTHREEbpbzFIVEONETWO$ & $\stsFOURnhTHREEbpbzgen$/5 \\
$A_5$ & 3 & $\staFIVEnhTHREEbppTHREETWO$ & $\staFIVEnhTHREEbppFIVEONETWO$ & $\staFIVEnhTHREEbmpTHREETWO$ & $\staFIVEnhTHREEbmpFIVEONETWO$ & --- & --- \\
$A_5$ & 4 & $\staFIVEnhFOURbppTHREETWO$ & $\staFIVEnhFOURbppFIVEONETWO$ & $\staFIVEnhFOURbmpTHREETWO$ & $\staFIVEnhFOURbmpFIVEONETWO$ & $\staFIVEnhFOURbpbzFIVEONETWO$ & $\staFIVEnhFOURbpbzgen$/5 \\
$S_5$ & 3 & $\stsFIVEnhTHREEbppTHREETWO$ & $\stsFIVEnhTHREEbppFIVEONETWO$ & $\stsFIVEnhTHREEbmpTHREETWO$ & $\stsFIVEnhTHREEbmpFIVEONETWO$ & --- & --- \\
$S_5$ & 4 & $\stsFIVEnhFOURbppTHREETWO$ & $\stsFIVEnhFOURbppFIVEONETWO$ & $\stsFIVEnhFOURbmpTHREETWO$ & $\stsFIVEnhFOURbmpFIVEONETWO$ & $\stsFIVEnhFOURbpbzFIVEONETWO$ & $\stsFIVEnhFOURbpbzgen$/5 \\
\bottomrule
\end{tabular}
\caption{Core grid (medians over 5 seeds; train $L{=}32$, test $L{=}512$; full per-cell ranges in the
artifacts). At and above the law width, $+b$ cells shortcut, $-b$ cells are exact; below it (rows
$S_4$@2, $A_5$@3, $S_5$@3) both fail to fit --- the necessity side of the law. Right columns: zeroing
$W_b$ at inference on the trained $+b$ model restores generalization at law-minimal widths.}
\label{tab:main}
\end{table}

\section{The law: minimal $n_h$ = generator reflection length, representation-relative}\label{sec:law}
\begin{figure}[t]
\centering\includegraphics[width=.62\linewidth]{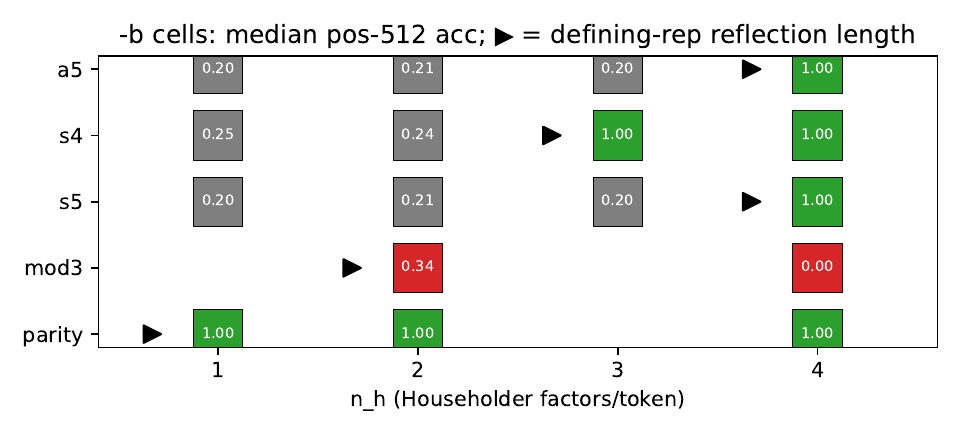}
\caption{$-b$ cells: median position-512 accuracy per (task, $n_h$); $\blacktriangleright$ marks the
defining-representation generator reflection length. Generalization begins exactly at the law width
(green), below it nothing fits (gray); mod-3 (drift, \S\ref{sec:mech}) is the labeled exception.}
\label{fig:law}
\end{figure}

For a permutation generator $P$, its reflection length in the defining representation is
$\operatorname{rank}(I-P)$: a transposition is 1 reflection, a 4-cycle 3, a 5-cycle 4. Because each
step must realize a generator as a product of $n_h$ Householder factors, representability requires
$n_h \ge \max_g \operatorname{rank}(I-P_g)$ \emph{in whatever faithful representation the model
realizes}. Empirically (Fig.~\ref{fig:law}), the $-b$ models obey this law exactly in its
defining-representation form: minimal generalizing $n_h$ = $\stminnhparity$ (parity),
$\stminnhsFOUR$ ($S_4$), $\stminnhaFIVE$ ($A_5$), $\stminnhsFIVE$ ($S_5$), and every below-law cell is
no-fit --- e.g.\ $S_5$ at $n_h{=}3$: $\stsFIVEnhTHREEbmpTHREETWO$ at position 32. For $S_5$ the
necessity is representation-independent: in \emph{every} faithful representation of $S_5$ the
5-cycle's reflection length is 4 (checked across the irreducible representations), so $n_h{=}4$ is a
true lower bound, which the $n_h{=}3$ cell confirms behaviorally.

For $S_4$ and $A_5$, however, \citet{siems2025deltaproduct} report robust extrapolation at
$n_h{=}2$ via their isomorphisms to rotation groups in $SO(3)$, where every element is a product of
$\le 2$ reflections. Our $S_4$/$A_5$ cells at $n_h{=}2$ (and $A_5$ at 3) are no-fit --- \emph{both}
results are correct, and the reconciliation is the point: their format (group-element classification
from a matrix state) admits the compact $SO(3)$ realization; our format (point tracking through a
vector state with a per-step linear readout) pins the defining permutation representation, e.g.\
because the $SO(3)$ realization of $S_4$ carries the four objects as unsigned diagonal axes whose
sign ambiguity a linear readout cannot resolve. The law that survives both datasets is
representation-relative: \emph{minimal per-step compute equals the generators' reflection length in
the cheapest faithful representation the task format admits}. Task format is thus a hidden variable
in state-tracking benchmarks --- two papers can disagree about ``the $n_h$ needed for $S_4$'' while
both being right.

\section{Mechanism: a parasitic attractor, discriminated by pre-registered arms}\label{sec:mech}
\begin{figure}[t]
\centering\includegraphics[width=.98\linewidth]{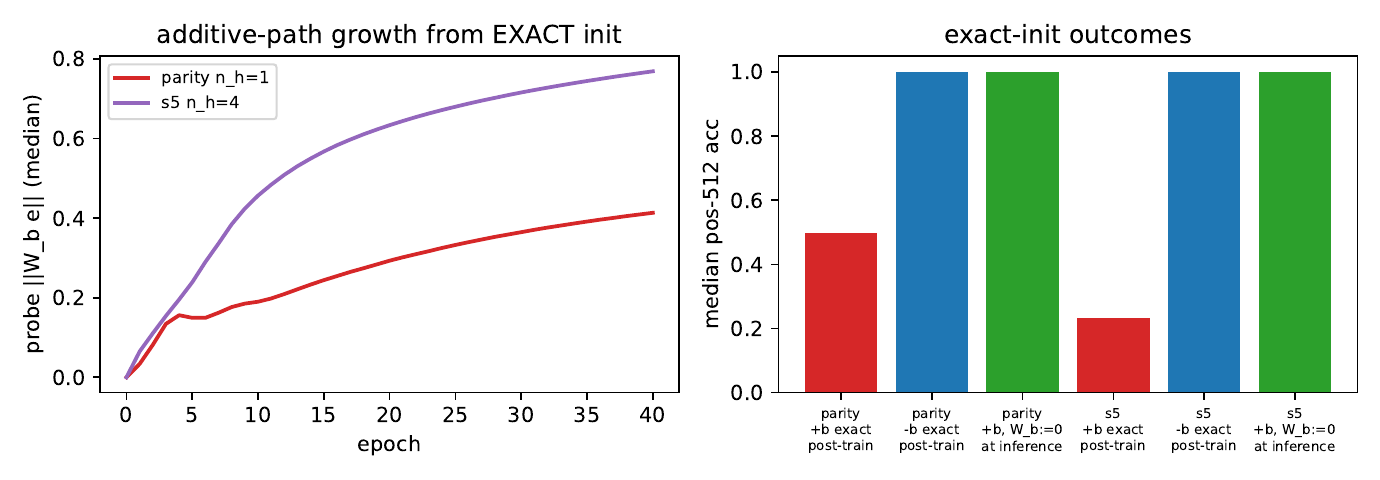}
\caption{E3, medians over 5 seeds. Left: from a verified-exact initialization with $W_b{=}0$,
training grows the additive path. Right: outcomes --- the $+b$ exact-init model is pulled off the
exact solution; the $-b$ control is stable; zeroing $W_b$ at inference restores it.}
\label{fig:exact}
\end{figure}

\textbf{The additive route attracts even from the exact solution (E3).} We construct exact solutions
by hand (reflection programs whose composed factors are asserted equal to the generator permutation
matrices; dead-axis factors implement identity), verify them (pre-training accuracy $1.00$ at every
probe position --- a registered instrument gate), set $W_b{=}0$, and train with the standard budget.
Across 5 seeds on both parity and $S_5$: the probe additive-path magnitude $\lVert W_b e\rVert$ grows
steadily from $0$ to $\stparitynhONEbpXusageend$ (parity) and $\stsFIVEnhFOURbpXusageend$ ($S_5$),
the state norm inflates (impossible through the norm-preserving reflections), and position-512
accuracy falls to $\stparitynhONEbpXpFIVEONETWO$/$\stsFIVEnhFOURbpXpFIVEONETWO$ --- while the $-b$
exact-init controls remain at $\stparitynhONEbmXpFIVEONETWO$/$\stsFIVEnhFOURbmXpFIVEONETWO$
(Fig.~\ref{fig:exact}). The additive pathway is not merely a basin that captures random
initializations; under Adam + cross-entropy it destabilizes the exact global solution itself. A
natural reading, stated as interpretation: orthogonal transitions cannot grow $\lVert h\rVert$, so
the additive path is the only channel through which cross-entropy's appetite for larger logits can
express itself, and what it adds is position-dependent drift that is harmless in-domain and fatal
out-of-distribution.

\textbf{Our registered mechanism prediction was wrong; the kill criterion fired; the truth is
better (P3/K2).} We pre-registered that the $+b$ shortcut's in-domain fit \emph{routes through} the
additive path (prediction: zeroing $W_b$ at inference collapses even position-32 accuracy). It does
not --- of the $\stbfitseeds$ seeds across all $+b$ cells that fit the training length,
$\stbfitsurvive$ retain in-domain accuracy under $W_b{:=}0$ (e.g.\ $S_5$:
$\stsFIVEnhFOURbpbzTHREETWO$ at position 32), so K2 fired as registered and the ``leaky-counter
\emph{route}'' story is withdrawn. What replaces it is stronger: at the law-minimal width, zeroing
$W_b$ at inference \emph{restores exact length generalization} --- parity
$\stparitynhONEbpbzFIVEONETWO$ ($\stparitynhONEbpbzgen$/5 seeds), $S_5$ $\stsFIVEnhFOURbpbzFIVEONETWO$
($\stsFIVEnhFOURbpbzgen$/5), $S_4$ $\stsFOURnhTHREEbpbzgen$/5, $A_5$ $\staFIVEnhFOURbpbzgen$/5. The
trained ``shortcut'' model at minimal width \emph{contains the exact automaton}: the reflections
converge to it anyway, and the additive path merely conceals it. The same holds for the exact-init
arms after being pulled off ($W_b{:=}0$ restores $\stparitynhONEbpXbzFIVEONETWO$/$\stsFIVEnhFOURbpXbzFIVEONETWO$).
The additive pathway is therefore \emph{parasitic} in the precise sense that it neither performs the
fit nor helps it; it grows because it is the cheapest confidence channel, and its growth both
destabilizes training away from exact solutions and masks exact solutions once learned.

\textbf{Boundaries (reported, not smoothed).} The concealment result is width- and task-scoped:
above the law width the freed reflection capacity co-adapts with the parasite and $W_b{:=}0$ restores
only $\stparitynhTWObpbzgen$--$\stparitynhFOURbpbzgen$/5 seeds on parity ($n_h{=}2,4$); per-seed
exceptions exist at law width ($S_4$ $\stsFOURnhTHREEbpbzgen$/5, $A_5$ $\staFIVEnhFOURbpbzgen$/5).
And mod-3 never locks: even $-b$, its median position-512 accuracy craters
($\stmodTHREEnhFOURbmpFIVEONETWO$ at $n_h{=}4$ --- systematic rotational drift, worse than chance,
with individual seeds at $1.00$). Notably this \emph{kills our own prior hypothesis} (from the June
pilot) that a transposition generator is what anchors exact solutions: $A_5$'s generators are all
even --- no transposition, no reflection among them --- yet $A_5$ locks exactly
($\staFIVEnhFOURbmpFIVEONETWO$). What separates the locking tasks from mod-3 is open; we note only
that mod-3 is abelian with a single continuously embeddable rotation generator, while the locking
tasks force non-commuting discrete constraints.

\textbf{Relation to rival mechanism accounts.} The unexplored-states hypothesis
\citep{buitrago2025length} attributes length-generalization failure to training-time state coverage;
our exact-init arms hold the initial solution (and hence its reachable state distribution) fixed and
still observe the pull-off, so state coverage is not the operative variable here. Error-control
theory \citep{chung2026error} proves affine trackers accumulate uncorrectable error at inference;
consistently, our $+b$ models fail OOD --- but the same weights with $W_b{:=}0$ are exact, and
training \emph{grows} $\bt$ from zero at an exact solution, so in this family the affine term is not
a fixed design constraint that error dynamics must cope with but a \emph{learned} liability that
optimization actively acquires. The two accounts are complementary to ours (inference-time and
data-distribution, respectively); the within-architecture ablation locates the training-time cause.

\section{Discussion}
\textbf{What this changes.} The expressivity-learnability gap in Householder linear RNNs has a
locatable, causal, single-term source in this setting --- and the failure it produces is not absence
of the solution but \emph{occlusion} of it. That reframes the practical question from ``how do we
make these models learn state tracking'' to ``how do we stop one pathway from first hijacking and
then hiding what they already learn.'' The minimal demonstration also yields an unreasonably cheap
diagnostic: evaluating a trained model with its additive input path zeroed is a one-line probe for a
concealed automaton, and at minimal width it doubles as a repair ($18/20$ seeds across the four group
tasks) --- with the honest caveat that per-seed exceptions exist and the trick degrades off the
minimal width.

\textbf{What this does not claim.} No universality: PD-SSM achieves exact $S_5$ extrapolation with an
additive term in a different architecture \citep{terzic2025pdssm}, and reveal-style supervision
makes the additive path load-bearing and helpful \citep{siems2026codestate}; our claims are scoped to
Householder-product linear RNNs under standard supervision, where the transitions alone are
expressive enough. No language-modeling prescription: in an LM the additive pathway carries content,
and these tasks use input only to select transitions; ``delete $W_b$'' is a probe and a mechanism
demonstration, not a training recipe. The law's necessity side is representation-independent only
where representation theory makes it so ($S_5$); elsewhere it is conditional on the format-pinned
representation, which is exactly the point.

\section{Limitations}
Single architecture family, optimizer, and budget; $d{=}112$; five tasks with two-generator
vocabularies; 5 seeds per cell (counts over $p$-values throughout); point-tracking format with
per-step linear readout. Pre-registration is repository-native (commit-stamped in our own history,
hashes in \S\ref{sec:setup} and Appendix~\ref{app:repro}), not third-party. The June pilot preceded
registration and is disclosed; its mod-3 ``flip'' did not survive 5 seeds (median craters; reported in
\S\ref{sec:mech}). The parasite account's scope beyond this family --- and what separates locking
tasks from drifting ones --- are open.

\subsection*{Acknowledgments}
This note answers, in a minimal system, the loss-landscape question posed by
\citet{shakerinava2026diagonal}, and builds directly on the DeltaProduct parameterization of
\citet{siems2025deltaproduct}.

\bibliographystyle{plainnat}
\bibliography{references}

\begin{thebibliography}{14}
\providecommand{\natexlab}[1]{#1}
\providecommand{\url}[1]{\texttt{#1}}
\expandafter\ifx\csname urlstyle\endcsname\relax
  \providecommand{\doi}[1]{doi: #1}\else
  \providecommand{\doi}{doi: \begingroup \urlstyle{rm}\Url}\fi

\bibitem[Chung et~al.(2026)Chung, Choi, and Kim]{chung2026error}
Jiwan Chung, Heechan Choi, and Seon~Joo Kim.
\newblock Rethinking state tracking in recurrent models through error control
  dynamics.
\newblock \emph{arXiv preprint arXiv:2605.07755}, 2026.

\bibitem[Del{\'e}tang et~al.(2023)Del{\'e}tang, Ruoss, Grau-Moya, Genewein,
  Wenliang, Catt, Cundy, Hutter, Legg, Veness, and Ortega]{deletang2023chomsky}
Gr{\'e}goire Del{\'e}tang, Anian Ruoss, Jordi Grau-Moya, Tim Genewein, Li~Kevin
  Wenliang, Elliot Catt, Chris Cundy, Marcus Hutter, Shane Legg, Joel Veness,
  and Pedro~A. Ortega.
\newblock Neural networks and the {Chomsky} hierarchy.
\newblock In \emph{International Conference on Learning Representations}, 2023.
\newblock arXiv:2207.02098.

\bibitem[Ebrahimi and Memisevic(2025)]{ebrahimi2025bilinear}
M.~Reza Ebrahimi and Roland Memisevic.
\newblock Revisiting bi-linear state transitions in recurrent neural networks.
\newblock In \emph{Advances in Neural Information Processing Systems}, 2025.
\newblock arXiv:2505.21749.

\bibitem[Grazzi et~al.(2025)Grazzi, Siems, Zela, Franke, Hutter, and
  Pontil]{grazzi2025negative}
Riccardo Grazzi, Julien Siems, Arber Zela, J{\"o}rg~K.H. Franke, Frank Hutter,
  and Massimiliano Pontil.
\newblock Unlocking state-tracking in linear {RNNs} through negative
  eigenvalues.
\newblock In \emph{International Conference on Learning Representations}, 2025.
\newblock arXiv:2411.12537.

\bibitem[Lee(2026)]{lee2026falsifier}
Jeonghoon Lee.
\newblock A held-out transition-pair falsifier for long-horizon non-abelian
  state tracking.
\newblock \emph{arXiv preprint arXiv:2606.07254}, 2026.

\bibitem[Merrill et~al.(2024)Merrill, Petty, and
  Sabharwal]{merrill2024illusion}
William Merrill, Jackson Petty, and Ashish Sabharwal.
\newblock The illusion of state in state-space models.
\newblock \emph{International Conference on Machine Learning}, 2024.
\newblock arXiv:2404.08819.

\bibitem[Mishra et~al.(2026)Mishra, Tan, Stoica, Gonzalez, and
  Dao]{mishra2026m2rnn}
Mayank Mishra, Shawn Tan, Ion Stoica, Joseph Gonzalez, and Tri Dao.
\newblock {M$^2$RNN}: Non-linear {RNNs} with matrix-valued states for scalable
  language modeling.
\newblock \emph{arXiv preprint arXiv:2603.14360}, 2026.

\bibitem[Ruiz and Gu(2025)]{buitrago2025length}
Ricardo~Buitrago Ruiz and Albert Gu.
\newblock Understanding and improving length generalization in recurrent
  models.
\newblock In \emph{Advances in Neural Information Processing Systems}, 2025.
\newblock arXiv:2507.02782.

\bibitem[Shakerinava et~al.(2026)Shakerinava, Khavari, Ravanbakhsh, and
  Chandar]{shakerinava2026diagonal}
Mehran Shakerinava, Behnoush Khavari, Siamak Ravanbakhsh, and Sarath Chandar.
\newblock The expressive limits of diagonal {SSMs} for state-tracking.
\newblock \emph{arXiv preprint arXiv:2603.01959}, 2026.

\bibitem[Siems et~al.(2025)Siems, Carstensen, Zela, Hutter, Pontil, and
  Grazzi]{siems2025deltaproduct}
Julien Siems, Timur Carstensen, Arber Zela, Frank Hutter, Massimiliano Pontil,
  and Riccardo Grazzi.
\newblock {DeltaProduct}: Improving state-tracking in linear {RNNs} via
  {Householder} products.
\newblock In \emph{Advances in Neural Information Processing Systems}, 2025.
\newblock arXiv:2502.10297.

\bibitem[Siems et~al.(2026)Siems, Grazzi, P{\"o}ppel, Kalinin, Ballani, and
  Rahmani]{siems2026codestate}
Julien Siems, Riccardo Grazzi, Korbinian P{\"o}ppel, Kirill Kalinin, Hitesh
  Ballani, and Babak Rahmani.
\newblock Learning state-tracking from code using linear {RNNs}.
\newblock \emph{arXiv preprint arXiv:2602.14814}, 2026.

\bibitem[Sung(2026)]{sung2026holonomic}
Ilmo Sung.
\newblock Robust reasoning as a symmetry-protected topological phase.
\newblock \emph{arXiv preprint arXiv:2601.05240}, 2026.

\bibitem[Terzi{\'c} et~al.(2025{\natexlab{a}})Terzi{\'c}, Hersche,
  Camposampiero, Hofmann, Sebastian, and Rahimi]{terzic2025sdssm}
Aleksandar Terzi{\'c}, Michael Hersche, Giacomo Camposampiero, Thomas Hofmann,
  Abu Sebastian, and Abbas Rahimi.
\newblock On the expressiveness and length generalization of selective
  state-space models on regular languages.
\newblock In \emph{AAAI Conference on Artificial Intelligence},
  2025{\natexlab{a}}.
\newblock arXiv:2412.19350.

\bibitem[Terzi{\'c} et~al.(2025{\natexlab{b}})Terzi{\'c}, Menet, Hersche,
  Hofmann, and Rahimi]{terzic2025pdssm}
Aleksandar Terzi{\'c}, Nicolas Menet, Michael Hersche, Thomas Hofmann, and
  Abbas Rahimi.
\newblock Structured sparse transition matrices to enable state tracking in
  state-space models.
\newblock In \emph{Advances in Neural Information Processing Systems},
  2025{\natexlab{b}}.
\newblock arXiv:2509.22284.

\end{thebibliography}

\appendix
\section{Reproducibility}\label{app:repro}
Pre-registration (grid, predictions P-R0/E1/E2/E3/P3, kill criteria K0--K2, decision rules) is commit
\texttt{2cf13e5}, made before any grid artifact existed; verdicts against it, unchanged, are commit
\texttt{84e21c8}. The grid is 202 runs (\texttt{statetrack\_note/runs/}, one JSON artifact each,
git-hash-stamped): 5 seeds per primary cell, 3 for references. Runner
(\texttt{statetrack\_note/runner.py}) asserts group closures, generator reflection lengths, and
exact-construction correctness programmatically before running; it is idempotent. Every number above
is a macro in \texttt{numbers.tex} generated by \texttt{analyze.py} from the artifacts and
byte-verified by \texttt{verify\_regen.py}. Model/training configuration: $d{=}112$, embeddings and
readout linear, 40 epochs, batch 512, $n{=}4000$ sequences of length 32, Adam ($3{\times}10^{-3}$),
per-position cross-entropy; evaluation $n{=}1000$ at $L{=}512$. Hardware: one RTX 3080; full grid
$\approx$ 53 minutes.
\end{document}